\documentclass[letterpaper, 10 pt, conference]{ieeeconf}  

\IEEEoverridecommandlockouts                              

\usepackage{graphicx}
\usepackage{amsmath}
\usepackage{amssymb}
\usepackage{CJKutf8}
\usepackage{xurl}
\usepackage{tabularx}
\usepackage{booktabs, tabularx, multirow}

\title{\LARGE \bf
Realistic Lip Motion Generation Based on 3D Dynamic Viseme and Coarticulation Modeling for Human-Robot Interaction 
}

\author{Sheng Li, Jingcheng Huang, and Min Li*
\thanks{*The authors are with The State Key Laboratory of Intelligent Manufacturing Equipment and Technology, Huazhong University of Science and Technology, Wuhan 430074, China.
        Min Li is the corresponding author.
        {\tt\small \{m202570574, D202380251, min.li\}@hust.edu.cn}}%
}

\begin{document}

\maketitle
\thispagestyle{empty}
\pagestyle{empty}

\begin{abstract}

Realistic lip synchronization is essential for the natural human-robot non-verbal interaction of humanoid robots. Motivated by this need, this paper presents a lip motion generation framework based on 3D dynamic viseme and coarticulation modeling. By analyzing Chinese pronunciation theory, a 3D dynamic viseme library is constructed based on the ARKit standard, which offers coherent prior trajectories of lips. To resolve motion conflicts within continuous speech streams, a coarticulation mechanism is developed by incorporating initial-final (Shengmu-Yunmu) decoupling and energy modulation. After developing a strategy to retarget high-dimensional spatial lip motion to a 14-DOF lip actuation system of a humanoid head platform, the efficiency and accuracy of the proposed architecture is experimentally validated and demonstrated with quantitative ablation experiments using the metrics of the Pearson Correlation Coefficient (PCC) and the Mean Absolute Jerk (MAJ). This research offers a lightweight, efficient, and highly practical paradigm for the speech-driven lip motion generation of humanoid robots. The 3D dynamic viseme library and real-world deployment videos are available at \url{https://github.com/yuesheng21/Phoneme-to-Lip-14DOF}

\end{abstract}

\section{INTRODUCTION}

As humanoid robots are increasingly deployed in real-world applications, multimodal human-robot interaction (HRI) is attracting growing attention \cite{1}\cite{2}. To ensure natural human-robot interaction (HRI) experiences, intuitive non-verbal behaviors alongside intelligent conversational capabilities are desired for humanoid robots \cite{3}\cite{4}. In particular, the synchronization of lip movements with speech is crucial for mitigating interaction dissonance. Psychological studies indicate that human speech perception is inherently a process of audiovisual multimodal fusion. The McGurk \cite{5} effect demonstrates that the dynamics of visual articulation directly reshape auditory signal decoding in the brain. Consequently, a spatiotemporal mismatch between lip movements and speech disrupts user immersion and triggers cognitive conflict, thereby exacerbating the uncanny valley effect \cite{6}.

Meanwhile, high-fidelity lip synchronization provides substantial engineering and humanistic value in complex environments. The dynamics of visual lips could significantly improve speech intelligibility under severe noise interference, which is providing a gain equivalent to an increase of approximately 15 dB in the signal-to-noise ratio (SNR) \cite{7}. This capability enables robots to convey information efficiently in noisy settings. Furthermore, precise lip movements serve as a critical assistance  for hearing-impaired and elderly individuals, as visual cues yield a comprehension improvement of 20\% to 60\% for these demographics \cite{8}. Therefore, equipping humanoid robots with high-fidelity lip synchronization capabilities significantly improves communication efficiency under challenging acoustic conditions.

Recent research on the lips motion generation for humanoid robots has advanced significantly, generally falling into two primary paradigms. The first category encompasses traditional rule\_based methods \cite{9}\cite{10}\cite{11}. Which typically construct a phoneme-to-viseme mapping to convert recognized speech units into predefined targets for mouth shapes, subsequently generating continuous sequences of facial motion through linear or polynomial interpolation. The second category involves data-driven generative models characterized by end-to-end deep learning and imitation, such as Wav2Lip \cite{12} and KMTalker \cite{13}. which mainly utilize large-scale audio and video datasets and generative adversarial networks (GANs) or Transformer architectures to directly learn the nonlinear mapping of audio features to two-dimensional (2D) videos. The extracted 2D movements of facial features are then retargeted to the mechanical degrees of freedom(DoF) of the robot \cite{14}\cite{15} . However, both paradigms exhibit limitations in the performance of real-world deployment.

Specifically, traditional methods of viseme modeling exhibit two inherent limitations. First, the static characters of these models cause a loss of dynamic features \cite{16}. To reduce computational complexity, traditional approaches typically reduce visemes to a 2D static snapshot, neglecting the dynamics of speech and 3D features. Robots exhibit discontinuous movements when performing lip motion generation. Second, the approaches developed based on English phonetics present obstacles in cross-lingual adaptation, particularly when transferring to Chinese. Complex Chinese syllables involve multi-stage continuous sliding motions. Chinese exhibits a prominent co-articulation effect \cite{17}, where the movements of articulatory organs between adjacent phonemes influence each other. The rigid application of traditional static rules often leads to motion discontinuity and the loss of intermediate visemes, thereby diminishing the naturalness of the interaction.

\begin{figure*}[t]
    \centering
    \includegraphics[width=1.0\textwidth]{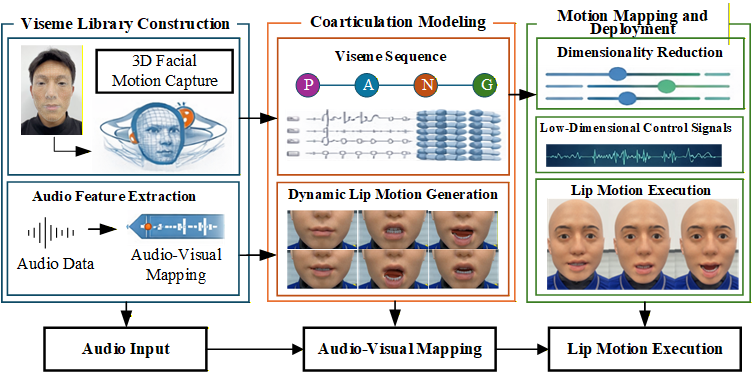} 
    \caption{ Framework  of lip motion generation}
    \label{fig:sys_frame}
\end{figure*}

Meanwhile end-to-end models face dual challenges involving a dimensional gap and computational barriers during deployment in the real world. Retargeting 2D motion of facial features to the robots with multiple DoFs necessitates complex mappings of inverse kinematics, which frequently lose spatial depth and details of high-frequency deformation. The black-box characteristics and inherent lack of kinematic controllability prevent engineers from fine-tuning low-level parameters to correct errors in execution. Furthermore, a severe reliance on computational resources poses a critical practical limitation. High-dimensional generative models depend heavily on external computing hardware, hindering real-time inference on devices with constrained computation. The resulting delays directly disrupt the real-time performance required for voice interaction.

To overcome the aforementioned limitations, this paper proposes a lightweight dynamic lip motion generation and coarticulation modeling method for real-world HRI in Chinese. The main contributions of this paper are as follows:

(1) A 3D dynamic viseme library is established. By constructing a library comprising 14 3D dynamic viseme sequences to serve as the underlying engine, this method overcomes the constraints of traditional 2D static mapping and reproduces the nonlinear motion trajectories of human lips during articulation. 

(2) Addressing the unique articulatory mechanisms of the Chinese language, this paper designs a fusion algorithm based on initial-final decoupling, which effectively resolves the challenges of coarticulation and endows lip motions with human-like dynamic tension. 

(3) A dimensionality reduction for motion mapping and a hybrid calibration mechanism are formulated to translate high-dimensional facial features  into low-dimensional motion control of a robotic platforms, completing the pathway from algorithmic development to real-world execution.

\section{METHODOLOGY}

To achieve high-fidelity lip synchronization, a lip motion generation framework based on 3D dynamic viseme and coarticulation modeling is developed as illustrated in Fig. 1, which includes dynamic viseme library construction, coarticulation modeling, and motion mapping and deployment. Specifically, driven by audio input, the system maps the text sequence recognized from the audio to the 3D dynamic viseme library to generate an initial viseme sequence. Subsequently, kinematic fusion on complex speech flows to resolve motion conflicts between adjacent visemes is performed using viseme coarticulation modeling. Finally, dynamic lip motions are generated and deployed on a humanoid robot using motion mapping methods. This process completes the pathway from digital motion generation to hardware deployment. The theoretical design and implementation details are elaborated in the subsequent sections.

\subsection{Dynamic Viseme Library Construction}

To reproduce high-fidelity articulatory details, a 3D dynamic viseme library is constructed, which is different from traditional approaches that treat visemes as 2D discrete static snapshots. As the core data asset of the system, this library precomputes complex nonlinear facial movements into searchable 3D trajectory sequences through offline recording and refined processing.

\noindent \textbf{1) Spatiotemporal definition of 3D dynamic viseme}

Spatiotemporally, the variation of mouth shape during articulation constitutes a continuous biomechanical process encompassing muscle contraction, maintenance, and relaxation. Traditional modeling approaches using single-frame target points discard the kinematic and micro-motion information in this process. To finely decouple local facial movements, this research constructs a parameterized space utilizing the ARKit standard \cite{18}, comprising 52 facial blendshapes. Given the highly localized characteristics of speech-driven motions, this approach extracts 27 core bases strongly correlated with lip movements, such as \textit{jawOpen}, regulating jaw amplitude and \textit{mouthFunnel} controlling the lip rounding. The 3D trajectories of the lip feature points for each viseme are defined as a spatiotemporal sequence incorporating the temporal variations of these 27 bases. This representation records not only the deformation endpoints but also the complete 3D dynamic paths.

\noindent \textbf{2)Viseme mapping based on Chinese pronunciation theory}

Visemes are classified based on the phonological structure of Chinese, specifically focusing on Initials (consonant-like components) and Finals (vowel-centered components). This distinction is crucial as it guides the decoupling of lip motions in our co-articulation model. The Chinese phonological system comprises 21 initials and 39 finals \cite{19}, generating hundreds of articulatory combinations. According to phonetic theory, variations in tongue placement primarily determine auditory differences, whereas lip shape and jaw aperture dictate the visual representation. Studies demonstrate that auditorily distinct phonemes frequently exhibit high visual similarity, a phenomenon termed homophones \cite{20}.

To eliminate visual redundancy, this method performs a many-to-one dimensionality reduction mapping on over 60 pinyin combinations based on kinematic features, merging phonemes sharing identical morphological and dynamic behaviors. Consequently, a library comprising 14 categories of core dynamic visemes (detailed in Table I) is constructed to achieve precise phoneme-to-viseme mapping. This framework adheres to the visual clustering principles of the MPEG-4 FBA standard \cite{21} while deeply adapting to the articulatory characteristics of Chinese, including complex medials and initial-final coarticulation. Fig. 2 illustrates the spatiotemporal evolution of the articulation cycle for the typical viseme 'V1\_BPM' alongside the ARKit \textit{jawOpen}, and for the viseme 'V2\_F' alongside the \textit{mouthUpperUp} (averaging the \textit{mouthUpperUpLeft} and \textit{mouthUpperUpRight}). This visualization highlights the explicit correspondence between lip motions and the associated numerical parameters.

\begin{table}[t]
\centering
\caption{Classification of Chinese Visemes and Their Corresponding Phonetic Syllables} 
\label{table_viseme_mapping}
\renewcommand{\arraystretch}{1.2} 
\footnotesize 
\setlength{\tabcolsep}{2pt} 

\begin{tabularx}{\columnwidth}{l X X X} 
\toprule 
\textbf{Viseme ID} & \textbf{Description} & \textbf{Key Features} & \textbf{Mapped Syllables} \\
\midrule 

V1\_BPM & Bilabial & Closed-mouth & \textit{b, p, m} \\
V2\_F   & Labiodental & Lip-biting & \textit{f} \\
V3\_D   & Apical & Slightly-spread & \textit{d, t, n, l} \\
V4\_GKH & Velar & Neutral-open & \textit{g, k, h} \\
V5\_JQX & Palatal & Unrounded-open & \textit{j, q, x, y} \\
V6\_ZCS & Dental sibilant & Teeth-bared & \textit{z, c, s} \\
V7\_ZH  & Retroflex & Slightly-pursed & \textit{zh, ch, sh, r, er} \\

V8\_A   & Wide-open & Wide-open & \textit{a, ua, an} \\
V9\_O   & Rounded & Mid-rounded & \textit{o, uo, ou} \\
V10\_E  & Spread & Mid-spread & \textit{e, en, eng} \\
V11\_I  & Teeth-aligned & Tight-spread & \textit{i, in, ing} \\
V12\_U  & Closed-mouth & Tight-rounded & \textit{w, u} \\
V13\_V  & Pursed-lip & Funneled & \textit{ü, v, ue, iu, iong} \\
V14\_AI & Semi-open & Semi-open-rounded & \textit{ai, ei, an, ui} \\

\bottomrule 
\end{tabularx}
\end{table}


\begin{figure}[t]
    \centering
    \includegraphics[width=0.5\textwidth]{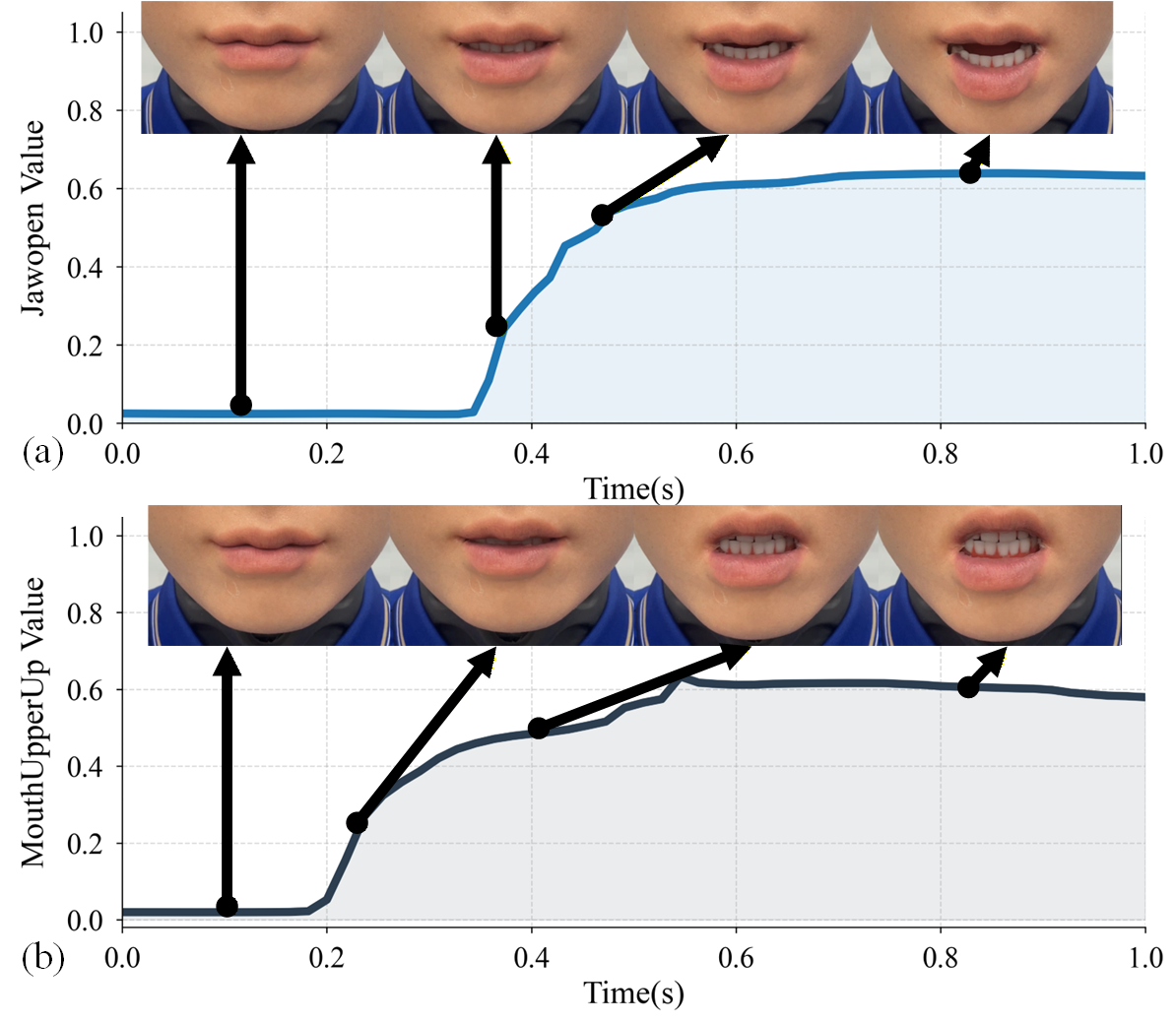} 
    \caption{ Lip shapes and  \textit{JawOpen/MouthUpperUp} plots of two typical visemes, (a) V1\_BPM, b) V2\_F}
    \label{fig:viseme_time}
\end{figure}

\noindent \textbf{3) Dynamic viseme library construction}

\begin{figure}[t]
    \centering
    \includegraphics[width=0.5\textwidth]{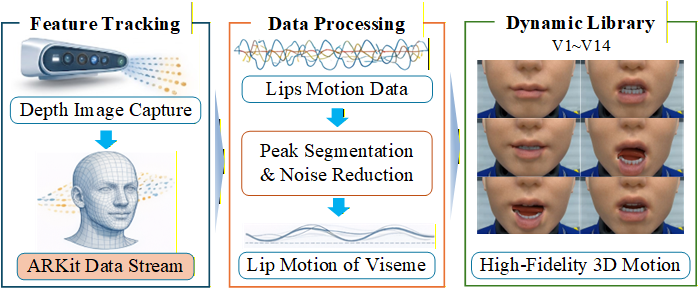} 
    \caption{Flowchart of viseme library construction}
    \label{fig:make_dynamic_viseme}
\end{figure}

The flowchart for a high-precision dynamic viseme library construction is illustrated in Fig. 3, which comprises of facial feature tracking, viseme extraction and normalization. Initially, the depth sensing technology is introduced to track facial features highly correlated with lip muscle movements during continuous pronunciation with ARKit standard. After collecting the periodic lip motion generated by the continuous pronunciation, the high-frequency noise is filtered out, and the continuous signal stream is segmented at its peaks according to kinematic periodic patterns to extract the complete motion envelope of a single pronunciation.  Then, the dynamic temporal alignment and mean fusion are applied to the multiple extracted pronunciation cycle trajectories, fitting a high-fidelity standard motion curve for the target viseme. After extracting all target visemes, a global duration normalization was performed on the standard trajectories along the time axis, ultimately constructing a dynamic viseme feature space with high spatiotemporal consistency.

\subsection{Lip Motion Generation Based on Coarticulation Modeling}

The articulation with a period \textit{T} is derived from the timestamped text sequence extracted during preprocessing, which facilitates the retrieval of a base lip motion sequence from the dynamic viseme library. While simple linear concatenation of these sequences generates continuous movements, it fails to reproduce the complex coarticulation effects inherent in human speech. For example, during the articulation of the syllable ‘tu’, the lips already exhibit the lip-rounding tendency of the final /u/ at the exact moment the initial /t/ is produced. To simulate this physiological characteristic, a hierarchical motion generation strategy is proposed based on initial-final decoupling and temporal coarticulation, successfully achieving smooth transitions across the lip states of the robot.

Chinese characters are converted to Pinyin representations and decomposed into underlying viseme sequences with lengths of one to three. To process viseme sequences with varying lengths, corresponding nonlinear temporal generation mechanisms are designed. For pure monophthongs (such as /a/) or compound vowels with pre-recorded complete dynamic features (such as /ei/), the pronunciation remains free from the interference of other visemes. The system directly extracts the complete spatiotemporal trajectory from the corresponding viseme library and outputs it without any fusion intervention. For the cases with two and three visemes, the details are expressed as follows.

\noindent \textbf{1) Dual-viseme fusion}




Addressing the most prevalent initial-final combinations, such as the syllable `ba' comprising the initial /b/ and the final /a/, this approach introduces a dynamic fusion weight $w$ that varies continuously with time $t$. $\tau = t/T \in [0, 1]$ denotes the normalized temporal progress. The facial motion represented by viseme state vector $\mathbf{V}_{blend}$ is formulated as follows:

\begin{equation}
    \mathbf{V}_{blend}(\tau) = (1 - w(\tau, a)) \cdot \mathbf{V}_1(\tau) + w(\tau, a) \cdot \mathbf{V}_2(\tau)
\end{equation}

\begin{equation}
    w(\tau, a) = \left( \frac{1 - \cos(\pi \tau)}{2} \right)^a
\end{equation}

$\mathbf{V}_1(\tau)$ and $\mathbf{V}_2(\tau)$ denote the 27-dimensional lip blendshape state vectors corresponding to the initial and the final at time $\tau$, respectively. When $w$ approaches 0, the facial motion is dominated by the initial, whereas it is dominated by the final as it approaches 1. Acoustically, the initial typically occupies a brief fraction of the total duration, generally falling below 20\% \cite{22}. 
If constant-velocity linear blending is strictly enforced based on the real acoustic timing during physical robot control, the underlying physical actuators lack adequate response time to execute the complete strokes of critical actions, such as lip closure and lip biting. This limitation induces severe visual phoneme omission and high-frequency vibrations. To balance the brevity of acoustic articulation with the physical completeness of mechanical motions, a nonlinear cosine function incorporating an exponential bias $a$ is designed for the weight $w$. While a standard cosine curve guarantees smooth velocity transitions at the onset and termination of the movement, this biased function artificially prolongs the initial retention period of the control weight for the initial, by configuring $a=0.7$. This modification effectively simulates the acoustic property where the initial accounts for a minimal time proportion.

\noindent \textbf{2)Three-viseme fusion}

For common combinations comprising an initial and a compound final (e.g., the syllable ``zuo'', which consists of the initial /z/ and the finals /u/ and /o/), this complex, extended articulation period is divided into two continuous transitional sub-states along the temporal axis. Let $\mathbf{V}_1$ (initial, /z/), $\mathbf{V}_2$ (the first final, /u/), and $\mathbf{V}_3$ (the second final, /o/) denote the three sequential viseme states deconstructed from the given syllable. Within the normalized articulation cycle $\tau \in [0, 1]$, a temporal division ratio $\lambda$ is introduced to partition the motion generation process into two distinct stages. The ultimately generated viseme state vector $\mathbf{V}_{blend}(\tau)$ is formulated as the following piecewise function:


{\small 
\begin{equation}
\scalebox{0.95}{
$
\mathbf{V_{blend}}(\tau) =
\begin{cases}
\left[ 1 - w\left( \frac{\tau}{\lambda}, 1 \right) \right] \mathbf{V_1} + w\left( \textstyle\frac{\tau}{\lambda}, 1 \right) \mathbf{V_2} \\ 
\quad (0 \leq \tau \leq \lambda) \\[2ex] 
\left[ 1 - w\left( \frac{\tau - \lambda}{1 - \lambda}, 1 \right) \right] \mathbf{V_2} + w\left( \textstyle\frac{\tau - \lambda}{1 - \lambda}, 1 \right) \mathbf{V_3}\\ 
\quad (\lambda < \tau \leq 1)
\end{cases}
$
}
\tag{3}
\end{equation}
} 


The first segment of this function characterizes the transition from the initial to the first final, whereas the second segment delineates the progression from the first final to the second final. This cascaded mechanism effectively decouples the complex high-dimensional spatiotemporal trajectory into two distinct phases. This decoupling minimizes the risk of omitting intermediate visemes and endows the system with nuanced Chinese prosodic expressiveness.

\subsection{Motion Mapping and Deployment}

Constrained by internal space and mechanical complexity, there are limited numbers of independent DOFs for the robots. This structural limitation precludes a simple one-to-one mapping between the 27-dimensional digital commands and $K(<27)$ DOFs for the robots. For instance, within the ARKit digital feature space, \textit{mouthSmileLeft} and \textit{mouthStretchLeft} represent entirely independent feature dimensions. However, according to the mechanical topology, the execution of both actions must multiplex the same left-mouth-corner traction actuator. 

To bridge this software-hardware mapping discrepancy, this research designs a sparse linear combination mapping layer. This layer linearly superimposes multiple digital motion commands assigned to the same actuator with specific weight coefficients, thereby calculating the final target stroke of the specific actuator. This dimensionality reduction mapping process is formulated as follows:

\begin{equation}
    \mathbf{M}(t) = [\mathbf{W}] \cdot \mathbf{V}(t) \tag{4}
\end{equation}

Where, $\mathbf{V}(t) \in \mathbb{R}^{27 \times 1}$ denotes the smoothed 27-dimensional digital blendshape vector at time $t$, and $\mathbf{M}(t) \in \mathbb{R}^{K \times 1}$ represents the resulting control vector for the robots. The core mapping matrix $[\mathbf{W}] \in \mathbb{R}^{K \times 27}$ defines the energy distribution weights from the high-dimensional blendshape space to the low-dimensional motion space.

To determine the weight parameters of the mapping matrix $[\mathbf{W}]$, a hybrid calibration mechanism that integrates data-driven motion capture with expert-guided heuristic refinement is adopted. Initially, a vision-based facial motion capture system acquires benchmark data of human facial movements during articulation to construct 27-dimensional digital blendshapes. By comparing with the baseline physical response curves of the robot, a fundamental linear mapping matrix is established, which establishes a bionically grounded coarse prior for $[\mathbf{W}]$. 

Given the highly nonlinear deformation characteristics inherent in the stretching of the silicone facial skin and the inevitable physical tolerances of the mechanical linkages, a purely mathematical linear mapping frequently fails to achieve the desired anthropomorphic realism. Consequently, relying on the established motion capture priors, high-frequency dynamic lip motion tests are conducted on the physical robot. Using visual similarity and motion naturalness as evaluation metrics, targeted manual compensation is applied to key non-zero weight nodes within the matrix.

Through this hybrid calibration paradigm, the system effectively compensates for the nonlinear physical errors introduced by flexible materials. This process endows the robot with highly expressive bionic lip dynamics, completing the pathway from digital generation algorithms to physical hardware actuation.

\section{EXPERIMENTAL ANALYSIS AND DISCUSSION}

The proposed method for lip motion generation is deployed on the humanoid robot head for experimental validation and demonstration

\subsection{Experimental Preparation} 

Before conducting experiments, the robot platform, dynamic viseme library, and testing corpus are prepared.

\noindent \textbf{1) Experimental setup}

As illustrated in Fig. 4, a high-fidelity humanoid robot head with a total of 29 DoFs is utilized as the validation platform, featuring a hybrid actuation system through linkage mechanisms and cable drives. The actuators simulate the contraction and relaxation of human facial muscles by pulling anchor points located on the inner surface of the flexible silicone skin. To address the lip generation task, the oral region of this platform is densely configured with 14 degrees of freedom. This structural design provides exceptional kinematic flexibility, enabling the reproduction of a full spectrum of lip dynamics, encompassing closure, rounding, spreading, and jaw aperture variations as shown in Fig. 4b.

\begin{figure}[t]
    \centering
    \includegraphics[width=0.5\textwidth]{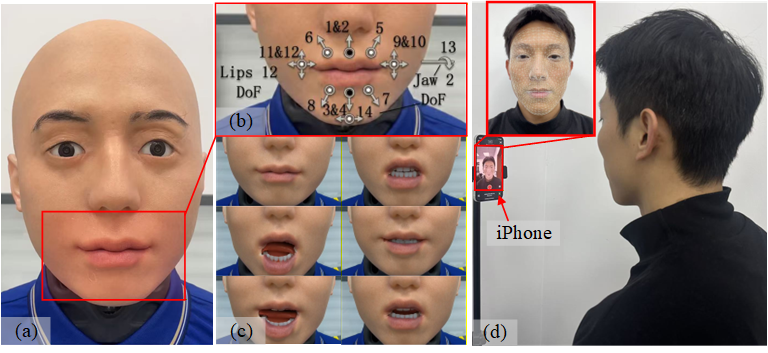} 
    \caption{Experimental setup. (a) humanoid robot platform, (b) DoFs of the lips and jaw, (c) lip-syncing demonstration, (d) Data acquisition setup.}
    \label{fig:exp_robot}
\end{figure}

\noindent \textbf{2) Dynamic viseme library construction}

As illustrated in Fig.4d an iPhone equipped with a TrueDepth camera system is utilized as the high-precision data collection terminal, which integrates an infrared projector, an infrared camera, and a flood illuminator to construct a 3D topology by projecting over 30,000 invisible infrared dots onto the face. It also coordinates with the Live Link Face data transmission protocol for data synchronization. During the collection process, subjects repeatedly pronounce specific target phonemes, such as the consonant /b/. The system analyzes the subtle movements of facial muscles in real time through the ARKit framework. At a constant sampling rate of 60 frames per second (FPS), the device outputs a real-time data stream containing 52 sets of standard blendshape coefficients. 27 motion parameters directly related to lip morphology from this stream are extracted and recorded. With the procedures illustrated in Fig. 2, a 3D dynamic viseme library is constructed, which provides solid data support for subsequent animation synthesis and driving experiments.

\noindent \textbf{3)  Testing corpus preparation}

A testing corpus capable of providing comprehensive viseme coverage is introduced for comprehensive evaluation. This dataset comprises four representative Chinese short sentences, with each sentence containing five characters as shown in Table II. The corpus selection follows the principles of maximizing phoneme diversity and kinematic difficulty, aiming to cover the pronunciation and lip shapes of all key initials and finals in Mandarin. The test set covers highly challenging pronunciation scenarios. For instance, S1 focuses on the coarticulation of compound vowels, which requires smooth handling of the three-stage continuous deformation from slightly open to rounded lips and then to a wide opening during the pronunciation of /gua/. S3 contains numerous bilabial consonants, such as /b/, /p/, and /m/, serving to verify the capability of complete closure between the upper and lower lips to resolve the visual phoneme omission issue observed in baseline methods. S4 tests the high-frequency dynamic response of the system during rapid transitions between unrounded visemes and rounded visemes.

\begin{table}[t]
\centering
\caption{Testing corpus} 
\label{table_test_corpus}
\renewcommand{\arraystretch}{1.5} 
\footnotesize 
\setlength{\tabcolsep}{2pt} 

\begin{tabularx}{\columnwidth}{@{} c l X @{}} 
\toprule
\textbf{ID} & \textbf{Test Sentences (with Pinyin)} & \textbf{Key Phonetic Features} \\
\midrule

S1 & \begin{CJK*}{UTF8}{gbsn}一个大西瓜\end{CJK*} (Yi Ge Da Xi Gua) &  Transition of compound finals \\
S2 & \begin{CJK*}{UTF8}{gbsn}师傅喝绿茶\end{CJK*} (Shi Fu He L\"u Cha) & Rounded and pursed-lip visemes \\
S3 & \begin{CJK*}{UTF8}{gbsn}爸爸买白菜\end{CJK*} (Ba Ba Mai Bai Cai) & Bilabial closures (b, p, m) \\
S4 & \begin{CJK*}{UTF8}{gbsn}自己做早餐\end{CJK*} (Zi Ji Zuo Zao Can) & Spread apicals and rounded visemes \\

\bottomrule
\end{tabularx}
\end{table}

\subsection{Experimental Design}

To comprehensively evaluate the effectiveness of the proposed framework and the contributions of each core module, ablation experiments are designed with four comparable strategies.

\textbf{Method A:} Functions as a traditional static baseline approach. It exclusively extracts static viseme target points and relies on simple linear interpolation to generate transition trajectories.

\textbf{Method B:} A dynamic viseme direct-drive approach. This strategy incorporates the dynamic viseme while applying a hard temporal segmentation at phoneme boundaries.

\textbf{Method C:} Functions as the fundamental coarticulation approach. It incorporates an initial-final coarticulation fusion algorithm based on a perceptual library.

\textbf{Method D:} Extending Method C, it further integrates dynamic amplitude modulation based on audio RMS energy and a moving average filtering mechanism.

Facial feature motion and synchronized video are collected from human subjects when reading the four aforementioned testing sets. The clean audios are extracted from the video recordings as the input. Ultimately, four sets of 27-dimensional blendshape coefficient temporal trajectories with a sampling rate of 60 FPS are acquired, establishing the foundation for subsequent quantitative evaluations and anthropomorphism analyses.

\subsection{Experimental Results and Discussion  }

To validate and demonstrate the performance of the proposed framework, the smoothness, accuracy and anthropomorphic naturalness of the lip motions are evaluated.

\begin{figure}[t]
    \centering
    \includegraphics[width=0.5\textwidth]{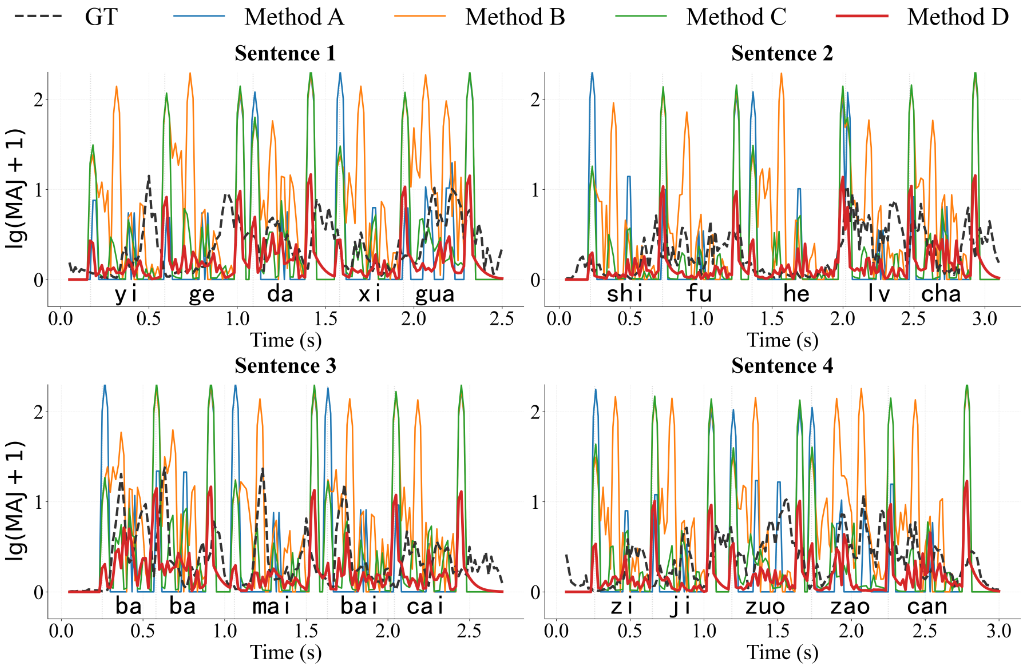} 
    \caption{MAJ Plots of lip motion generated by 4 methods}
    \label{fig:jerk}
\end{figure}



\begin{table}[t]
\caption{Experimental results of average jerk}
\label{table_average_jerk}
\centering
\begin{tabular}{lccccc}
\toprule
\multicolumn{6}{c}{\textit{Average Jerk ($10^3 /s^3$)} $\downarrow$} \\
\midrule
Driver Method & S1 & S2 & S3 & S4 & Average \\
\midrule
Groundtruth & 1.91 & 1.32 & 2.18 & 1.70 & 1.78 \\
Method A & 11.98 & 9.40 & 15.12 & 10.01 & 11.63 \\
Method B & 25.63 & 15.26 & 21.21 & 21.27 & 20.84 \\
Method C & 13.04 & 9.72 & 13.69 & 10.64 & 11.77 \\
\textbf{Method D} & \textbf{1.07} & \textbf{0.86} & \textbf{1.20} & \textbf{0.90} & \textbf{1.01} \\
\bottomrule
\end{tabular}
\end{table}

\noindent \textbf{1)	Smoothness of lip motions}

The parameter \textit{jerk} (the derivative of acceleration) is introduced as a motion smoothness metric to reflect motion abruptness and impact. Excessive jerking causes overheating, unnatural jitter, or mechanical damage. 

Although all 27 blendshape coefficients are associated with the lip and jaw, certain coefficients, some of them (such as \textit{JawLeft}, \textit{JawRight}, and \textit{MouthDimple}) are minimally activated during natural pronunciation. Therefore, this study extracts the nine most active key blendshape coefficients, which exhibit significantly higher variances. The mean absolute jerk (MAJ) is calculated and compared to quantify the smoothness cost across four methods with real human facial data introduced as the ground truth (GT) as illustrated in Fig. 6 and Table III. Some findings can be observed:

\begin{itemize}
    \item[--] Method B generates a high average jerk peak of 20.84 due to the lack of a phoneme transition algorithm, inducing severe discontinuities.
    \item[--] Although Method C implements coarticulation, its reliance on dynamic data interpolation introduces more fluctuations than the single-frame targets of Method A. Consequently, Method C yields a slightly higher jerk than Method A, with both exhibiting substantial mutations.
    \item[--] Method D demonstrates exceptional smoothness; its average jerk of 1.01 is marginally lower than the 1.78 of human motion capture data. This occurs because empirical capture inevitably incorporates muscle micro-tremors and marker noise.
\end{itemize}

Therefore, the proposed method filters microscopic jitter while preserving macroscopic amplitude, supplying actuators with a secure kinematic trajectory.

\noindent \textbf{2)Accuracy and anthropomorphic naturalness}

To objectively evaluate the accuracy and anthropomorphic naturalness of the lip motions, the Pearson correlation coefficient (PCC) \cite{23} and the root mean square error (RMSE) are introduced as the evaluation metrics. The PCC evaluates the morphological similarity between the generated motion and the human motion. This metric reflects the synchronization capability of the algorithm regarding temporal phase, articulatory rhythm, and fluctuation trends, collectively representing temporal synchrony. The RMSE penalizes deviations of the generated motion, reflecting the precision of the physical opening and closing scales, which represent the spatial amplitude deviation.  High PCC and low RMSE could demonstrate a highly anthropomorphic naturalness of the motion generation algorithm.

The jaw joint functions as the reference node for lower facial movements, establishing the physical spatial framework for lip morphological changes. The degree of jaw opening directly determines the baseline tension and 3D global coordinates of the perioral muscles. Additionally, the periodic opening and closing of the jaw share a distinct physical mapping with the short-time energy envelope of the speech signal and the formants of core vowels. Since the jaw is the articulatory organ with the largest displacement amplitude and highest visual salience, even a slight temporal phase shift or amplitude overshoot in its trajectory immediately disrupts the naturalness of lip-audio synchronization. Thereby, the temporal \textit{JawOpen} coefficient serves as the decisive metric for assessing the accuracy and anthropomorphic naturalness. The \textit{JawOpen} plots of lip motion generated by 4 methods are compared with the GT in Fig. 6, and the corresponding PCC and RMSE are shown in Tables VI and V. The execution results using the S1 sentence are subsequently presented and compared with the human articulatory motions as illustrated in Fig. 7. Some findings can be observed:

\begin{figure}[t]
    \centering
    \includegraphics[width=0.5\textwidth]{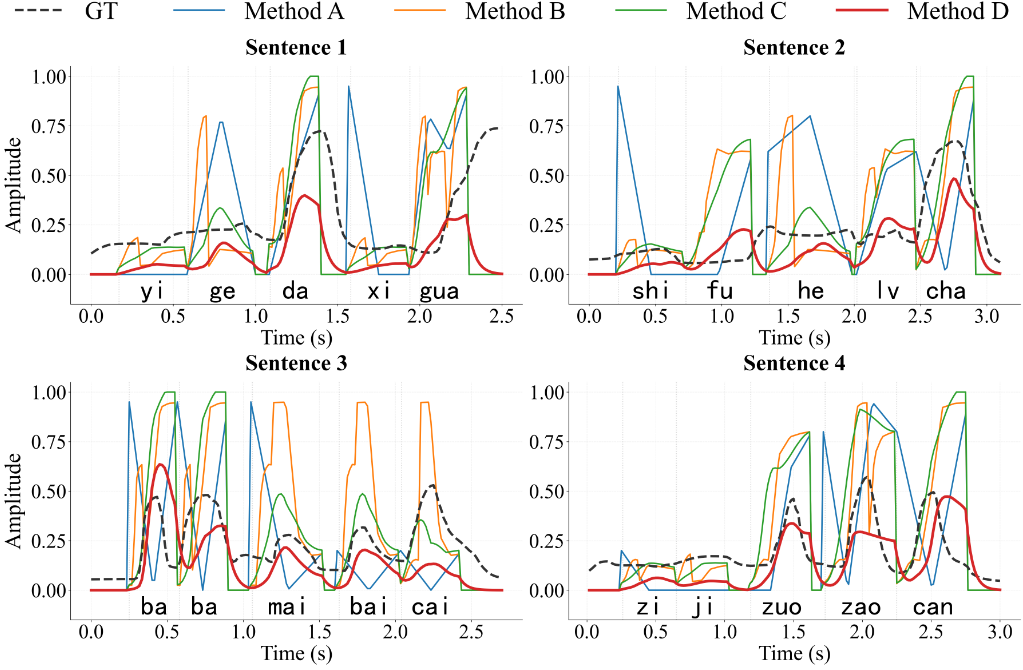} 
    \caption{\textit{JawOpen} plot of lip motion generated by 4 methods}
    \label{fig:jwaopen}
\end{figure}

\begin{table}[t]
\centering
\caption{PCC and RMSE of lip motion generated by 4 methods}
\label{table_quantitative_comparison}
\renewcommand{\arraystretch}{1.2} 
\footnotesize 
\setlength{\tabcolsep}{4pt} 

\begin{tabularx}{\columnwidth}{@{} l X X X X X @{}} 
\toprule
\textbf{Methods} & \textbf{S1} & \textbf{S2} & \textbf{S3} & \textbf{S4} & \textbf{Average} \\
\midrule

\multicolumn{6}{l}{\textit{Trajectory Similarity (PCC $\uparrow$)}} \\ \addlinespace[0.5ex]
Method A & 0.112 & 0.191 & -0.028 & 0.347 & 0.155 \\
Method B & 0.216 & 0.382 & 0.532 & 0.562 & 0.443 \\
Method C & 0.295 & 0.477 & 0.588 & 0.595 & 0.489 \\
\textbf{Method D} & \textbf{0.505} & \textbf{0.649} & \textbf{0.602} & \textbf{0.624} & \textbf{0.595} \\

\midrule 

\multicolumn{6}{l}{\textit{Trajectory Accuracy (RMSE $\downarrow$)}} \\ \addlinespace[0.5ex]
Method A & 0.352 & 0.316 & 0.298 & 0.294 & 0.315 \\
Method B & 0.321 & 0.283 & 0.323 & 0.292 & 0.305 \\
Method C & 0.304 & 0.255 & 0.251 & 0.309 & 0.280 \\
\textbf{Method D} & \textbf{0.256} & \textbf{0.147} & \textbf{0.165} & \textbf{0.139} & \textbf{0.177} \\

\bottomrule
\end{tabularx}
\end{table}

\begin{itemize}
    \item[--] The average PCC of Method A reaches merely 0.155, rendering it incapable of matching the human articulatory rhythm. Method B and Method C with coarticulation or dynamic transition mechanisms achieve average PCC of 0.443 and 0.489, respectively. However, as illustrated in Fig. 6, the absence of fine-grained amplitude modulation based on audio energy frequently causes these methods to exhibit excessive mouth opening. Consequently, their RMSEs remain substantially high at 0.305 and 0.280.
    
    \item[--] Benefiting from the synergistic effect of the energy mapping and the exponential moving average filtering mechanism, the average PCC of Method D escalates to 0.595, achieving the optimal rhythmic conformity across the four test sentences. Simultaneously, its average RMSE substantially decreases to 0.177, representing a 36.8\% reduction compared to Method C.
    
    \item[--] Despite its exceptional overall performance, Method D still exhibits localized trajectory deviations when processing complex syllables situated at the end of a sentence, such as the syllable `gua' in S1. A comparative analysis reveals that human articulation frequently entails a silent post-articulatory continuation movement driven by physical inertia after the acoustic signal terminates. Conversely, the kinematic state of the proposed method relies heavily on the continuous input of audio energy. This dependency causes the system to terminate the articulatory transition prematurely at the acoustic boundary, failing to completely replicate this visual lag phenomenon.
\end{itemize}

\begin{figure}[t]
    \centering
    \includegraphics[width=0.5\textwidth]{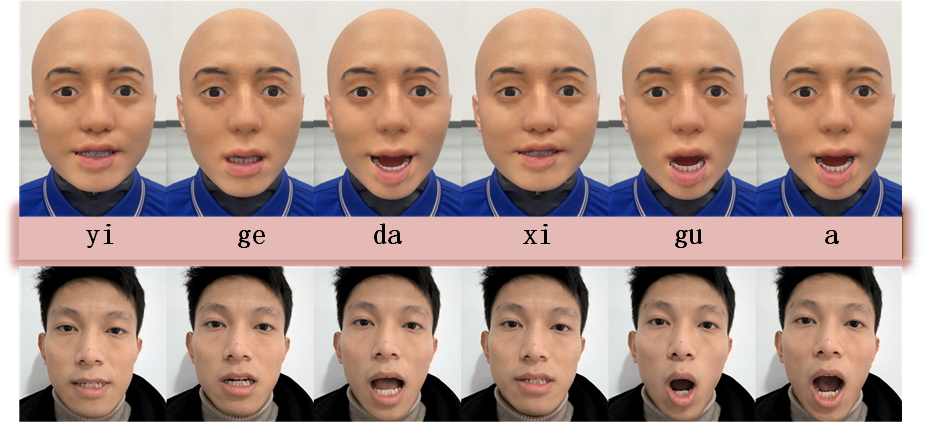} 
    \caption{Comparison of execution results using S1 sentence}
    \label{fig:my_label}
\end{figure}

\section{CONCLUSIONS}

This paper presents a lip motion generation framework based on 3D dynamic viseme and coarticulation modeling, addressing the challenges of natural human-robot non-verbal interaction. By constructing a 3D dynamic viseme library integrated with continuous 3D motions, the system overcomes the limitations of static 2D mapping and ensures the generation of natural speech rhythms through acoustic energy modulation. To facilitate the deployment, a dimensionality reduction for motion mapping mechanism accurately retargets high-dimensional digital blendshapes to the low-dimensional motion control of a robotic platform. Quantitative evaluations indicate that the proposed framework significantly improves the PCC and reduces the MAJ, providing high-fidelity lip motion under the constraints of edge computing.

While lip synchronization is effectively enhanced, the coordination of global facial expressions requires further investigation. Future research will explore the integration of large language models (LLMs) to establish an audio-semantic dual-modal drive for full facial coordination. Furthermore, the application of reinforcement learning will be investigated to enable the generalization of the framework across heterogeneous robotic platforms and multilingual interaction environments.

\section*{ACKNOWLEDGMENT}
This work was supported in part by the Fundamental and Interdisciplinary Disciplines Breakthrough Plan of the Ministry of Education of China under Grant JYB2025XDXM208, the National Natural Science Foundation of China under Grant 52575019 and 52188102, and the National Natural Science Foundation of China
Overseas Excellent Young Scholars Fund.

\addtolength{\textheight}{-10cm}   





\bibliographystyle{IEEEtran}
\bibliography{references}

@article{1,
  title={Intelligent Physical Robots in Health Care: Systematic Literature Review},
  author={Huang, Rong and Li, Hongxiu and Suomi, Reima and Li, Chenglong and Peltoniemi, Teijo},
  journal={Journal of Medical Internet Research},
  volume={25},
  pages={e39786},
  year={2023},
  publisher={JMIR Publications Inc., Toronto, Canada},
  doi={10.2196/39786}
}

@article{2,
  title={Customer experiences with service robots in hotels: a review and research agenda},
  author={Rana, Nripendra P and Begum, Nusaiba and Faisal, Mohd Nishat and Mishra, Anubhav},
  journal={Journal of Hospitality Marketing \& Management},
  volume={34},
  number={2},
  pages={145--174},
  year={2025},
  publisher={Taylor \& Francis},
  doi={10.1080/19368623.2024.2403640}
}

@article{3,
  title={Interactions with robots: The truths we reveal about ourselves},
  author={Broadbent, Elizabeth},
  journal={Annual Review of Psychology},
  volume={68},
  pages={627--652},
  year={2017},
  publisher={Annual Reviews},
  doi={10.1146/annurev-psych-010416-044058}
}

@article{4,
  title={Nonverbal Behavior of Service Robots in Social Interactions—A Survey on Recent Studies},
  author={Peter, Julia and Mertsching, B{\"a}rbel},
  journal={Applied Sciences},
  volume={13},
  number={17},
  pages={9882},
  year={2023},
  publisher={MDPI},
  doi={10.3390/app13179882}
}

@article{5,
  title={Hearing lips and seeing voices},
  author={McGurk, Harry and MacDonald, John},
  journal={Nature},
  volume={264},
  number={5588},
  pages={746--748},
  year={1976},
  publisher={Nature Publishing Group},
  doi={10.1038/264746a0}
}

@article{6,
  title={The Uncanny Valley},
  author={Mori, Masahiro},
  translator={MacDorman, Karl F and Kageki, Norri},
  journal={IEEE Robotics \& Automation Magazine},
  volume={19},
  number={2},
  pages={98--100},
  year={2012},
  publisher={IEEE},
  doi={10.1109/MRA.2012.2192811},
  note={(Original work published 1970)}
}

@article{7,
  title={Visual Contribution to Speech Intelligibility in Noise},
  author={Sumby, W. H. and Pollack, Irwin},
  journal={The Journal of the Acoustical Society of America},
  volume={26},
  number={2},
  pages={212--215},
  year={1954},
  publisher={Acoustical Society of America},
  doi={10.1121/1.1907309}
}

@article{8,
  title={Auditory and visual consonant recognition and audiovisual enhancement by adults with hearing loss},
  author={Thibodeau, Linda M},
  journal={The Journal of the Acoustical Society of America},
  volume={121},
  number={5},
  pages={3137--3137},
  year={2007},
  publisher={Acoustical Society of America},
  doi={10.1121/1.4781682}
}

@inproceedings{9,
  title={Design of Voice Synchronized Robotic Lips},
  author={Miyake, Kazuhiro and Takanobu, Hideaki and Shimoyama, Isao and Takanishi, Atsuo},
  booktitle={Proceedings of the 2007 IEEE International Conference on Robotics and Automation},
  pages={742--747},
  year={2007},
  organization={IEEE},
  doi={10.1109/ROBOT.2007.363825}
}

@inproceedings{10,
  title={Development of a Humanoid Reading System with Voice-Lip Synchronization},
  author={Takanishi, Atsuo and Takuma, Shunsuke and Takanobu, Hideki and Miyake, Kazuhiro},
  booktitle={Proceedings of the 2002 IEEE/RSJ International Conference on Intelligent Robots and Systems},
  volume={3},
  pages={2404--2409},
  year={2002},
  organization={IEEE},
  doi={10.1109/IRDS.2002.1041628}
}

@article{11,
  title={JALI: an animator-centric viseme model for expressive lip synchronization},
  author={Edwards, Pif and Landreth, Chris and Fiume, Eugene and Singh, Karan},
  journal={ACM Transactions on Graphics (TOG)},
  volume={35},
  number={4},
  pages={1--11},
  year={2016},
  publisher={ACM New York, NY, USA},
  doi={10.1145/2897824.2925984},
  note={SIGGRAPH 2016}
}

@inproceedings{12,
  title={A Lip Sync Expert Is All You Need for Speech to Lip Generation In the Wild},
  author={Prajwal, K R and Mukhopadhyay, Rudrabha and Namboodiri, Vinay P and Jawahar, C V},
  booktitle={Proceedings of the 28th ACM International Conference on Multimedia},
  pages={484--492},
  year={2020},
  doi={10.1145/3394171.3413531},
  note={Wav2Lip}
}

@inproceedings{13,
  title={KMTalk: Speech-Driven 3D Facial Animation with Key Motion Embedding},
  author={Xu, Zhihao and Gong, Shengjie and Tang, Jiapeng and Liang, Lingyu and Huang, Yining and Li, Haojie and Huang, Shuangping},
  booktitle={Proceedings of the IEEE/CVF Conference on Computer Vision and Pattern Recognition (CVPR)},
  pages={2172--2181},
  year={2024},
  doi={10.48550/arXiv.2409.01113}
}

@inproceedings{14,
  title={Learning realistic lip motions for humanoid face robots},
  author={Shamsuddin, Syamimi and Aljunid, Syed and Adnan, S and Yussof, H and Mohamed, S and Zahari, N and Kamat, S and Nordin, M},
  booktitle={2012 IEEE International Conference on Robotics and Biomimetics (ROBIO)},
  pages={1660--1665},
  year={2012},
  organization={IEEE},
  doi={10.1109/ROBIO.2012.6491204}
}

@inproceedings{15,
  title={SingingBot: An Avatar-Driven System for Robotic Face Singing Performance},
  author={Zhao, Siyuan and others},
  booktitle={2023 IEEE International Conference on Robotics and Biomimetics (ROBIO)},
  pages={1--6},
  year={2023},
  organization={IEEE},
  doi={10.1109/ROBIO58561.2023.10354761}
}

@inproceedings{16,
  title={Capture, Learning, and Synthesis of 3D Speaking Styles},
  author={Cudeiro, Daniel and Bolkart, Timo and Laidlaw, Cassidy and Ranjan, Anurag and Black, Michael J},
  booktitle={Proceedings of the IEEE/CVF Conference on Computer Vision and Pattern Recognition (CVPR)},
  pages={10101--10111},
  year={2019},
  doi={10.1109/CVPR.2019.01034}
}

@article{17,
  title={Audio-driven facial animation by joint end-to-end learning of pose and emotion},
  author={Karras, Tero and Aila, Timo and Laine, Samuli and Herva, Antti and Lehtinen, Jaakko},
  journal={ACM Transactions on Graphics (TOG)},
  volume={36},
  number={4},
  pages={1--12},
  year={2017},
  publisher={ACM New York, NY, USA},
  doi={10.1145/3072959.3073658},
  note={SIGGRAPH 2017}
}

@misc{18,
  title={{ARFaceAnchor.BlendShapeLocation}},
  author={{Apple Inc.}},
  howpublished={\url{https://developer.apple.com/documentation/arkit/arfaceanchor/blendshapelocation}},
  year={2017},
  note={[Accessed: 2-Mar-2026]}
}

@book{19,
  title={The phonology of standard Chinese},
  author={Duanmu, San},
  edition={2nd},
  year={2007},
  publisher={Oxford University Press}
}

@inproceedings{20,
  title={Decoding visemes: Improving machine lip-reading},
  author={Bear, Helen L and Harvey, Richard},
  booktitle={2014 IEEE International Conference on Acoustics, Speech and Signal Processing (ICASSP)},
  pages={3132--3136},
  year={2014},
  organization={IEEE},
  doi={10.1109/ICASSP.2014.6854177}
}

@article{21,
  title={Face and 2-D mesh animation in MPEG-4},
  author={Tekalp, A. Murat and Ostermann, J{\"o}rn},
  journal={Signal Processing: Image Communication},
  volume={15},
  number={4--5},
  pages={387--421},
  year={2000},
  publisher={Elsevier},
  doi={10.1016/S0923-5965(99)00055-7}
}

@incollection{22,
  title={Duration Study for the Bell Laboratories Mandarin Text-to-Speech System},
  author={Shih, Chilin and Ao, Benjamin},
  booktitle={Progress in Speech Synthesis},
  editor={van Santen, Jan P. H. and Sproat, Richard W. and Olive, Joseph P. and Hirschberg, Julia},
  pages={383--399},
  year={1997},
  publisher={Springer},
  address={New York, NY},
  doi={10.1007/978-1-4612-1894-4_24}
}

@incollection{23,
  title={Pearson correlation coefficient},
  author={Benesty, Jacob and Chen, Jingdong and Huang, Yiteng and Cohen, Israel},
  booktitle={Noise reduction in speech processing},
  pages={1--4},
  year={2009},
  publisher={Springer Berlin Heidelberg}
}

\end{document}